\documentclass[runningheads]{llncs}
\usepackage[T1]{fontenc}
\usepackage{amsmath}
\usepackage{amsfonts}
\usepackage{array}
\usepackage{cellspace}
\usepackage[multiple]{footmisc}
\usepackage{graphicx}
\usepackage{orcidlink}
\usepackage{booktabs}
\usepackage{dsfont}
\begin{document}
\title{EDGER: EDge-Guided with HEatmap Refinement for Generalizable Image Forgery Localization}
\titlerunning{EDGER: EDge-Guided with HEatmap Refinement for Generalizable IFL}

\newcommand{\footnoteEqCon}{\protect\footnote[1]{Both authors contributed equally to this research as co-first authors.}}
\newcommand{\footnotemarkEqCon}{\protect\footnotemark[1]{}}
\newcommand{\footnoteCorAuth}{\protect\footnote[2]{Corresponding author. Email: \email{dtle@selab.hcmus.edu.vn}}}
\newcommand{\footnotemarkCorAuth}{\protect\footnotemark[2]{}}

\author{
Minh-Khoa Le-Phan\footnotemarkEqCon\inst{1,2}\orcidlink{0009-0005-9707-4026}
\and
Minh-Hoang Le\footnoteEqCon\inst{1,2}\orcidlink{0009-0005-1501-8080}
\and \\
Minh-Triet Tran\inst{1,2}\orcidlink{0000-0003-3046-3041}
\and
Trong-Le Do\footnoteCorAuth\inst{1,2}\orcidlink{0000-0002-2906-0360}
}
\authorrunning{Minh-Khoa Le-Phan, Minh-Hoang Le, Minh-Triet Tran, and Trong-Le Do}
%

\institute{University of Science - VNU-HCM, Ho Chi Minh City, Vietnam \and
Vietnam National University, Ho Chi Minh City, Vietnam
\email{\{lpmkhoa22,lmhoang22\}@apcs.fitus.edu.vn}, \\
\email{tmtriet@fit.hcmus.edu.vn},
\email{dtle@selab.hcmus.edu.vn}
}

\maketitle
\begin{abstract}
\vspace{-7mm}
Text-guided inpainting has made image forgery increasingly realistic, challenging both SID and IFL. However, existing methods often struggle to point out suspicious signals across domains.
To address this problem, we propose \textbf{EDGER}, a patch-based, dual-branch framework that localizes manipulated regions in arbitrary resolution images without sacrificing native resolution.
The first branch, \textbf{Edge-Guided Segmentation}, introduces a \textbf{Frequency-based Edge Detector} to emphasize high-frequency inconsistencies at manipulation boundaries, and fine-tunes a SegFormer to fuse RGB and edge features for pixel-level masks.
Since edge evidence is most informative only when patches contain both authentic and manipulated pixels, we complement Edge-Guided Segmentation with a \textbf{Synthetic Heatmapping} branch, a classification-based localizer that fine-tunes a CLIP-ViT image encoder with LoRA to flag fully synthetic patches.
Together, Synthetic Heatmapping provides coarse, patch-level synthetic priors, while Edge-Guided Segmentation sharpens boundaries within partially manipulated patches, yielding comprehensive localization.
Evaluated in the MediaEval 2025, SynthIM challenge, Manipulated Region Localization Task's setting, our approach scales to multi-megapixel imagery and exhibits strong cross-domain generalization.
Extensive ablations highlight the complementary roles of frequency-based edge cues and patch-level synthetic priors in driving accurate, resolution-agnostic localization.

\vspace{-2mm}
\end{abstract}

\vspace{-10mm}
\section{Introduction}\label{sec:introduction}
\vspace{-3mm}

Text‐guided inpainting has become one of the most powerful tools for manipulating digital photographs. Modern generative models can regenerate missing regions by synthesizing realistic textures and semantic structures, making forgery almost indistinguishable from the original capture.
Current research divides forgery analysis into two subtasks: Synthetic Image Detection (SID) and Image Forgery Localization (IFL).
SID aims to predict whether an image is real or fully generated but ignores localizing manipulated content, undermining trust in the decision~\cite{spai,texturecrop,rine,dire}.
IFL requires pinpointing manipulated regions in partially modified images, yet existing methods are often tuned to a narrow source domain and rarely evaluated across generators, datasets, or manipulation pipelines~\cite{trufor,diff_forensics,hifi}. 
Moreover, when manipulated regions occupy only a small fraction of a large image, fixed-resolution methods~\cite{diff_forensics} have problems with downscaling high‑resolution images, while unscaled processing methods~\cite{trufor,hifi} are computationally expensive.
Given that generative models evolve rapidly, domain-specific methods degrade quickly. Consequently, there is a pressing need for cross-domain and generalized localization methods that remain reliable as generative models change.

To address these limitations, the Multimedia Evaluation (MediaEval) Benchmark 2025 introduced the SynthIM challenge, "Synthetic Images: Advancing detection of generative AI used in real-world online images"~\cite{mediaeval-synthim}, which comprises two tasks: \textbf{Real vs. Synthetic} and \textbf{Manipulated Region Localization}.
The latter task invites participants to detect synthetic content and localize manipulated regions even when images have undergone common degradations such as resizing, compression, or cropping.
Within this benchmark, the data are categorized into three groups: \textbf{Original (OR)}, \textbf{Spliced (SP)}, and \textbf{Fully Regenerated (FR)}.
Since FR images are generated end-to-end and exhibit generative noise across the entire image, defining a localized manipulated region is not straightforward. 
In contrast, SP images are produced by inpainting only a masked area, providing a well-defined localization target. 
Accordingly, our work focuses on detecting and localizing the manipulated regions in SP images.

Building on this setting, we introduce \textbf{EDGER}, a framework for generalizable image forgery localization. 
It is guided by two key observations. 
First, spliced content must be blended into the surrounding context, leaving boundary cues (e.g., smoothing or resampling) along the transition band. 
Second, manipulated regions differ from their surroundings at a region level, even when a patch contains no visible boundary. 
Motivated by these observations, EDGER adopts a patch-based, dual-branch framework for localizing manipulated regions in arbitrary resolution images.
Each image is divided into smaller patches, enabling analysis at native resolution without excessive memory usage.
For each patch, we introduce a \textbf{Frequency-based Edge Detector} that operates in the frequency domain to emphasize boundary cues between real and manipulated regions. This approach captures transition edges effectively while avoiding the heavy training cost of noise-residual models.
We next fine-tune a SegFormer model~\cite{segformer} to predict per-pixel manipulation masks, taking as input both the original patch and the extracted edge features. We refer to this component as the \textbf{Edge-Guided Segmentation (EGS)} branch.
Since edge cues are informative only when a patch contains both authentic and manipulated pixels, we add a \textbf{Synthetic Heatmapping (SH)} branch - a classification-based localizer obtained by LoRA~\cite{lora} fine-tuning a CLIP-ViT~\cite{clip} image encoder.
The SH branch labels patches as real or fully synthetic, while the EGS branch refines the boundaries within partially manipulated patches. Together, these two branches provide complementary strengths, achieving accurate and consistent localization across images of any resolution.

\vspace{1mm}
Our contributions are summarized as follows:
\vspace{-2mm}
\begin{itemize}
    \item We propose a patch-based, dual-branch framework for manipulation localization in arbitrary resolution images.
    \item We introduce a Frequency-based Edge Detector that emphasizes high-frequency inconsistencies, and an EGS branch by fine-tuning SegFormer to fuse RGB and edge features for pixel-level masks.
    \item For patches without boundary cues, we add an SH branch by LoRA fine-tuning a CLIP-ViT image encoder, so the model flags patches as fully synthetic or real.
    \item We conduct extensive experiments, including cross-dataset evaluation, to analyze the contribution of each component and demonstrate the effectiveness of our approach.
\end{itemize}


\vspace{-2mm}
\section{Related Work}\label{sec:related work}
\vspace{-2mm}

We briefly review prior work on \emph{Synthetic Image Detection (SID)} and \emph{Image Forgery Localization (IFL)}, followed by methods most relevant to our approach.  

\vspace{-1mm}
\subsection{Synthetic Image Detection (SID)}
Early efforts in SID focused on modeling distributional discrepancies between real and synthetic content. 
TextureCrop~\cite{texturecrop} selects informative patches to infer authenticity scores, while SPAI~\cite{spai} learns the distribution of real images and detects out-of-distribution samples across arbitrary resolutions. 
These methods are designed for image-level classification and typically overlook localized forgery cues, motivating complementary research in IFL.  

\vspace{-1mm}
\subsection{Image Forgery Localization (IFL)}
IFL methods aim to identify manipulated regions at pixel level.  
Noise-based approaches include Noiseprint~\cite{noiseprint}, which leverages camera-model-specific residuals for manipulation detection, and TruFor~\cite{trufor}, which extends this idea with Noiseprint++ trained on a larger camera set for stronger generalization.  
CAT-Net~\cite{catnet} introduces a dual-stream CNN that learns compression artifacts jointly from RGB and DCT domains, enabling robust splicing localization in both JPEG and non-JPEG images.  
MVSS-Net~\cite{mvss} proposes multi-view feature learning, jointly exploiting tampering boundary cues and noise residuals under multi-scale supervision, leading to improved cross-dataset generalization.  
HiFi~\cite{hifi} combines frequency and color features to predict masks that balance artifact sensitivity with image context, while Diff-Forensics~\cite{diff_forensics} improves edge alignment by adding an explicit boundary loss.  
More recently, IML-ViT~\cite{iml-vit} explores a pure Transformer architecture for forgery localization, showing that high-resolution capacity, multi-scale features, and explicit edge supervision enable strong performance despite limited data.  

Despite these advances, most IFL models are trained and evaluated within a limited source domain. Cross-domain robustness across generators, datasets, and manipulation pipelines remains an open challenge, which the SynthIM benchmark directly addresses.


\vspace{-5mm}
\section{Method}\label{sec:method}
\vspace{-4mm}

\begin{figure}
    \vspace{-2mm}
    \centering
    \includegraphics[width=1\linewidth]{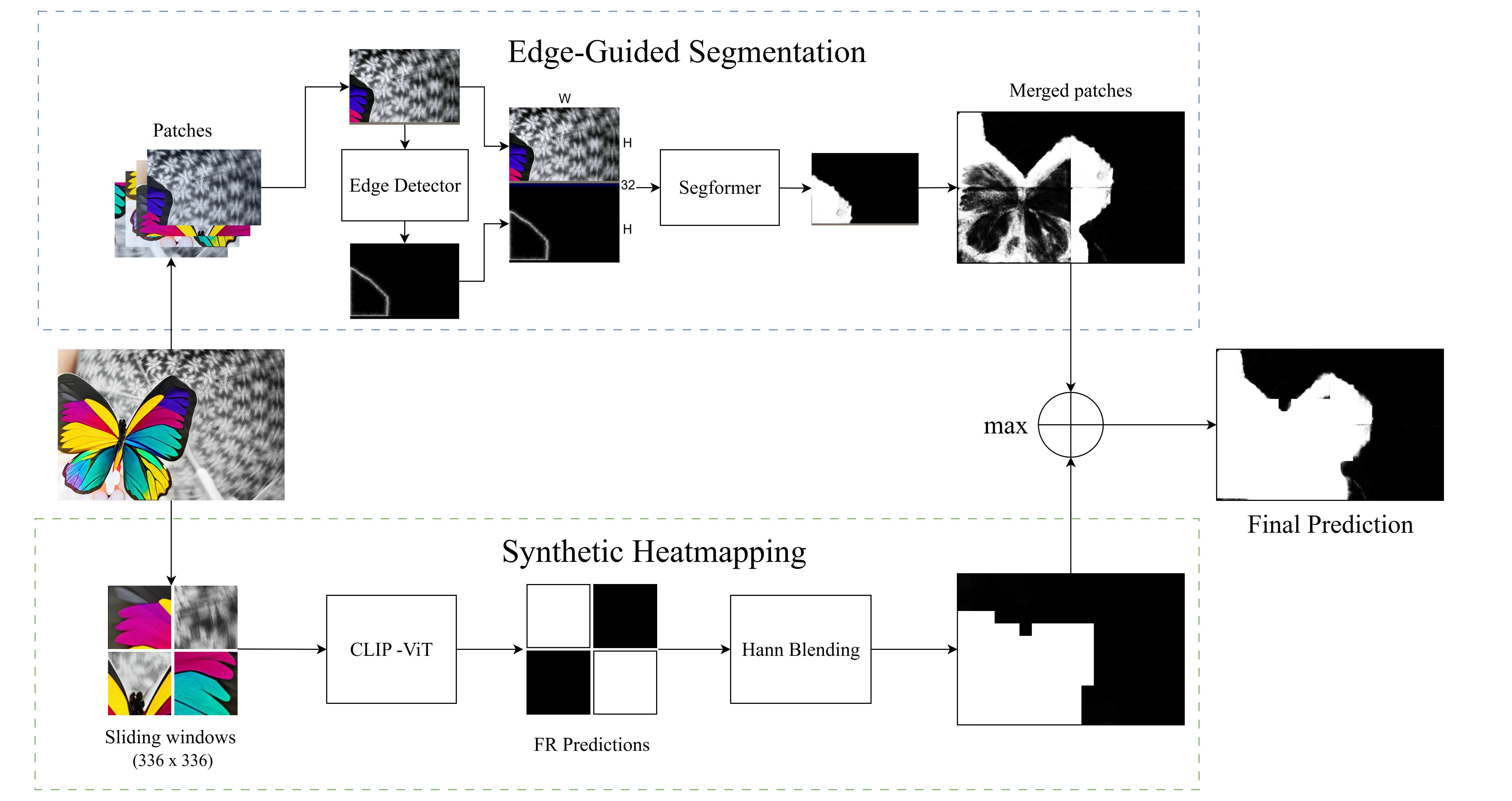}
    \vspace{-9mm}
    \caption{Overall localization pipeline. A patch-based dual-branch design: (i) the EGS branch injects a frequency-based edge prior (Section~\ref{sec:edge_detector}) into a SegFormer decoder; non-overlapping patches, padded to sizes divisible by 32, are stitched to full resolution. (ii) The SH branch applies a CLIP-ViT classifier over sliding windows with Hann blending to form a dense heatmap. The two outputs are fused by pixel-wise maximum, \(\mathbf{P}_{\mathrm{fuse}}=\max(\mathbf{P}_{\mathrm{EGS}},\mathbf{P}_{\mathrm{SH}})\), and thresholded to yield the final mask.}
    \label{fig:pipline-overall-localize}
    \vspace{-6mm}
\end{figure}

Our method targets the \textbf{Spliced (SP)} subset, where manipulated content is inserted via text-guided inpainting of a masked region, providing a clear localization target. It is guided by two key observations. \textbf{First}, spliced content must be blended into its context, leaving boundary cues (e.g., smoothing or resampling) along the transition band. \textbf{Second}, manipulated regions differ from their surroundings at a region level, even when no boundary is visible within a patch. To capture both cues, we propose \textbf{EDGER}, a patch-based dual-branch framework (Fig.~\ref{fig:pipline-overall-localize}). The \textbf{Edge-Guided Segmentation (EGS)} branch converts frequency-derived boundary cues into priors for mask prediction: a frequency-based edge detector (Section~\ref{sec:edge_detector}) produces an edge probability map that is injected into a SegFormer decoder~\cite{segformer}. Images are processed in non-overlapping patches and stitched into a full-resolution prediction. The \textbf{Synthetic Heatmapping (SH)} branch supplies region evidence: a CLIP-ViT classifier~\cite{clip}, fine-tuned for FR versus non-FR images, is applied in a sliding-window manner and blended with Hann weighting~\cite{hann-blending} to produce dense heatmaps that highlight synthetic areas even without boundaries. Finally, we fuse the two branches by pixel-wise max, \(\mathbf{P}_{\mathrm{fuse}}(u,v)=\max(\mathbf{P}_{\mathrm{EGS}}(u,v),\mathbf{P}_{\mathrm{SH}}(u,v))\), and threshold \(\mathbf{P}_{\mathrm{fuse}}\) to obtain the final mask.

\vspace{-4mm}
\subsection{Frequency-based Edge Detector}\label{sec:edge_detector}
\vspace{-1mm}

\begin{figure}
    \vspace{-5mm}
    \centering
    \includegraphics[width=1\linewidth]{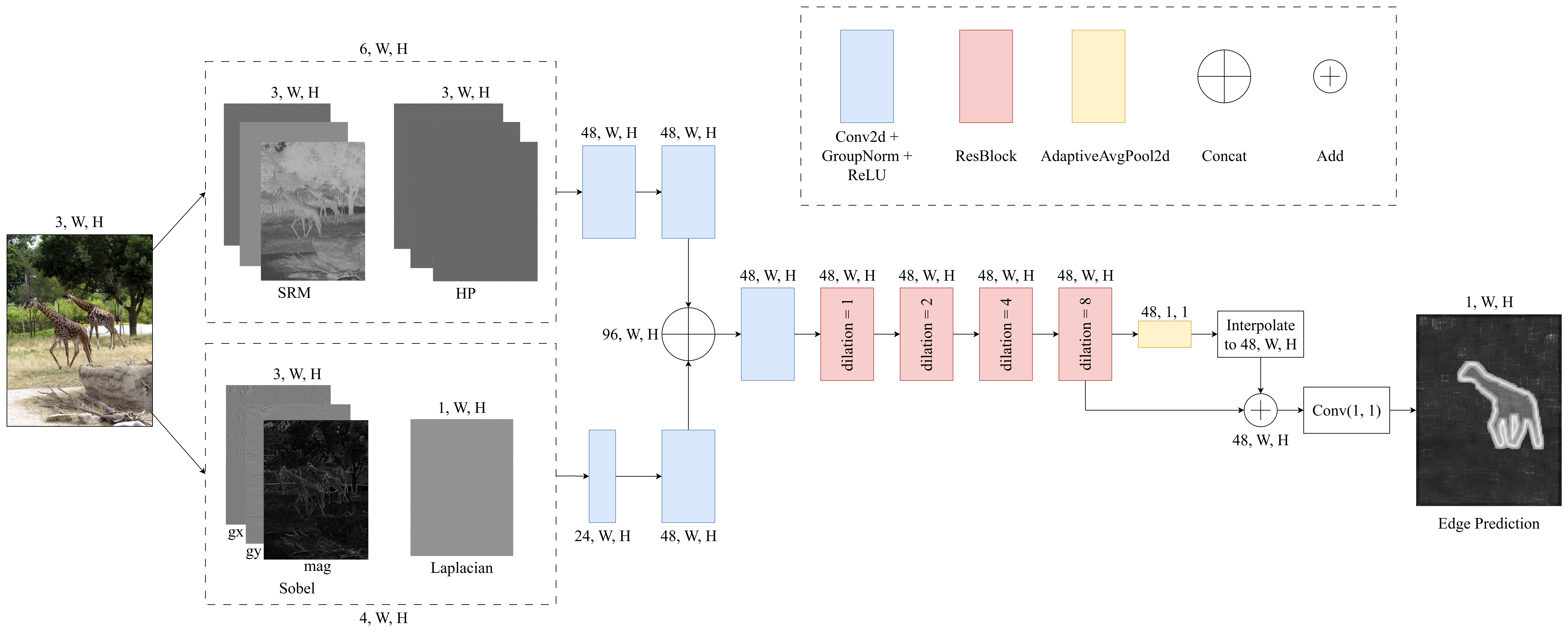}
    \vspace{-6mm}
    \caption{Pipeline of the Frequency-based Edge Detector. The network extracts high-pass and SRM-like residuals, computes fixed Sobel and Laplacian gradients from RGB, and fuses them through a multi-branch convolutional head composed of residual and dilated context blocks with global pooling. The head integrates frequency, gradient, and structural cues to predict soft edge probability maps supervised by multi-scale targets.}
    \label{fig:pipline-edge-detector}
    \vspace{-5mm}
\end{figure}

Manipulated boundaries often contain subtle inconsistencies that are easily suppressed by high-level feature extractors. To preserve these low-level signals, we design a frequency-based edge detector that produces a semantically agnostic edge probability map, illustrated in Fig.~\ref{fig:pipline-edge-detector}. Given an input image $\mathbf{I}\in\mathbb{R}^{3\times H\times W}$, we first derive residual features that expose high-frequency anomalies. A high-pass component is obtained as $\mathbf{R}_{\mathrm{hp}} = \mathbf{I} - \mathrm{Blur}_\sigma(\mathbf{I})$, while three SRM-like filters applied to the grayscale luminance capture prediction-error patterns. Concatenating the two gives a residual tensor $\mathbf{R}\in\mathbb{R}^{6\times H\times W}$. In parallel, we compute fixed gradients from the RGB image. Sobel operators yield horizontal, vertical, and magnitude responses, while a Laplacian kernel provides isotropic second-order structure, resulting in $\mathbf{F}_{\mathrm{grad}}\in\mathbb{R}^{4\times H\times W}$. These fixed filters are non-learnable and emphasize structural transitions introduced by splicing or inpainting. 

The residual and gradient tensors are then projected through shallow convolutional stems and fused. A lightweight stack of dilated residual blocks aggregates context at multiple receptive fields, while a global pooling branch introduces image-level priors. The fused representation is finally projected into a dense logit map $\mathbf{E}\in\mathbb{R}^{H\times W}$. For supervision, we avoid brittle binary edges and instead construct soft multi-scale edge targets. From a binary manipulation mask $\mathbf{M}$, morphological gradients at radii $s\in\mathcal{S}$ are smoothed with Gaussian kernels of $\sigma(s)=\lambda s$, then normalized:
\begin{equation}
\mathbf{Y}_{\mathrm{edge}} = \frac{\sum_{s\in\mathcal{S}} 
\mathrm{Gauss}_{\sigma(s)}\!\left(\mathrm{Dilate}(\mathbf{M};s) - \mathrm{Erode}(\mathbf{M};s)\right)}
{\max_{x,y}\sum_{s\in\mathcal{S}} \cdot + \varepsilon}.
\end{equation}
We use $\mathcal{S}=\{3,7,15\}$ by default. This produces soft transition bands around manipulated regions, encouraging stable predictions under annotation noise and post-processing. The resulting edge probability map serves as a reliable prior that is injected into the SegFormer decoder to sharpen localization boundaries.

\vspace{-2mm}
\subsection{Edge-Guided Segmentation (EGS) Branch}
\label{sec:egs}
\vspace{-1mm}

We integrate the frequency-based edge detector (Section~\ref{sec:edge_detector}) as a fixed prior for segmentation. The detector provides geometry-aware boundary cues that are independent of image semantics, while SegFormer supplies region-level context. Their combination sharpens contours and stabilizes mask predictions.

\begin{figure}
    \vspace{-6mm}
    \centering
    \includegraphics[width=0.8\linewidth]{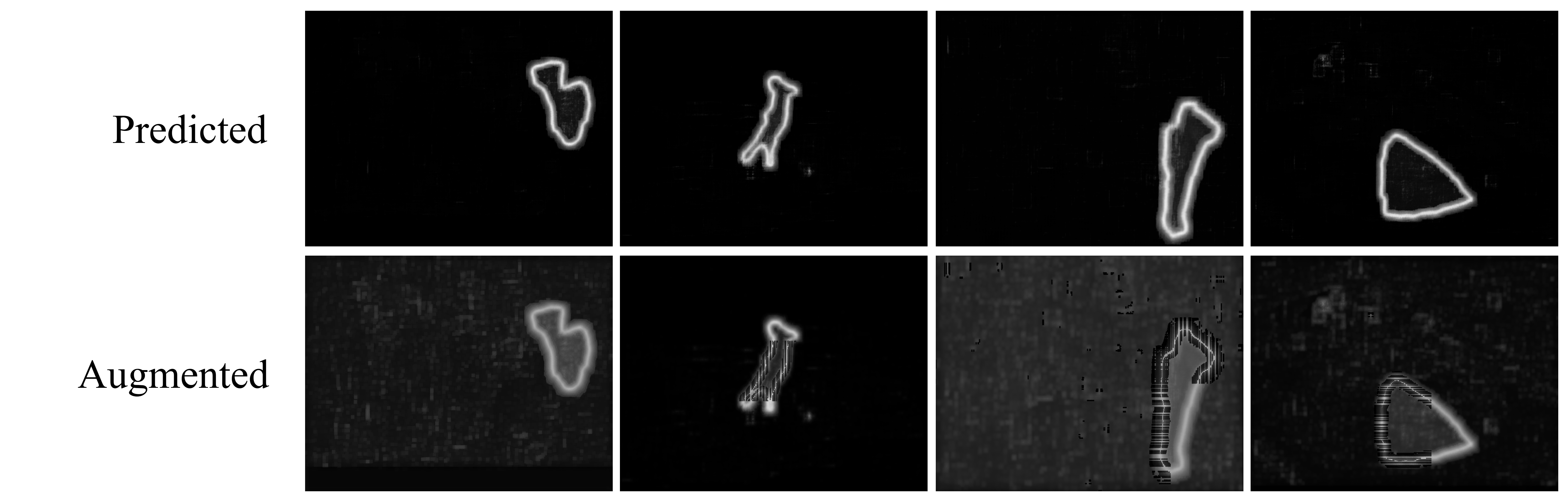}
    \vspace{-5mm}
    \caption{Predicted edge maps and their augmented counterparts used to simulate challenging test-time conditions. Augmentations lightly degrade or fragment boundaries (local background mixing, temperature softening, Gaussian blur, segment breaks, band-limited noise) to improve robustness.}
    \label{fig:nerf-edge}
    \vspace{-6mm}
\end{figure}

Formally, given an RGB image $\mathbf{X}\in\mathbb{R}^{3\times H\times W}$ and residual stack $\mathbf{R}\in\mathbb{R}^{C_r\times H\times W}$, the frozen edge head produces logits
\[
\mathbf{E}_{\ell} = f_{\mathrm{edge}}(\mathbf{X},\mathbf{R}), 
\qquad
\mathbf{E} = \sigma(\mathbf{E}_{\ell}) \in [0,1]^{1\times H\times W},
\]
which act as soft edge priors. To improve robustness, we optionally perturb $\mathbf{E}$ during training through a light augmentation $\mathcal{A}(\mathbf{E})$, mixing weak operators such as Gaussian blur, background confusion, and band-limited noise (see Figure~\ref{fig:nerf-edge}). At inference, we simply use $\tilde{\mathbf{E}}=\mathbf{E}$.

To inject the prior without altering SegFormer’s encoder interface, we adopt a spatial stacking strategy. Specifically, we build a “tall” three-channel canvas by vertically concatenating the original image $\mathbf{X}$, a separator band filled with per-image RGB means, and the replicated prior $\tilde{\mathbf{E}}$:
\[
\mathbf{X}_{\mathrm{tall}}
=
\begin{bmatrix}
\mathbf{X}\\
\mathbf{M}\\
\tilde{\mathbf{E}}\oplus\tilde{\mathbf{E}}\oplus\tilde{\mathbf{E}}
\end{bmatrix}
\in \mathbb{R}^{3\times H_2\times W}, \quad H_2=2H+s,
\]
where $\oplus$ denotes channel replication and $s$ ensures divisibility by 32. SegFormer processes $\mathbf{X}_{\mathrm{tall}}$ as a standard input, and the resulting logits are upsampled and cropped back to the original field of view:
\[
\mathbf{Z} = \mathrm{Crop}_{H,W}\Big(\mathrm{Up}\big(\mathrm{SegFormer}(\mathbf{X}_{\mathrm{tall}})\big)\Big),
\]

Finally, we expose $\mathbf{E}_{\ell}$ for optional auxiliary supervision with the soft multi-scale edge target (Section~\ref{sec:edge_detector}). The main segmentation loss is applied on $\mathbf{Z}$. Keeping the edge head frozen ensures the prior remains stable and data-agnostic, while the decoder learns to exploit boundary evidence without altering SegFormer’s backbone or positional encodings.

\vspace{-2mm}
\subsection{Synthetic Heatmapping (SH) Branch: Region Difference Detection}
\vspace{-1mm}

The Edge-Guided Segmentation (EGS) (Section~\ref{sec:egs}) branch enhances boundary sharpness, but it remains limited in scope. Boundary priors do not specify whether the interior or exterior of a contour is synthetic, leaving region-level decisions ambiguous. Patches without visible seams (fully real or fully generated) provide little structural evidence, making EGS prone to mislabeling and producing false masks after stitching. These issues become even more pronounced at ultra-high resolutions (e.g., 8K), where thousands of patches must be processed and aggregated, amplifying local mistakes and leaving small gaps after merging. To overcome these limitations, we introduce the Synthetic Heatmapping (SH) branch, illustrated in the lower branch of Fig.~\ref{fig:pipline-overall-localize}. This branch directly estimates region-level authenticity and complements the contour cues from EGS. The SH branch improves robustness both in ambiguous patches and across large-scale images.

\vspace{-5mm}
\subsubsection{Stage 1: Fully Regenerated (FR) Classification}
We first train a lightweight classifier to discriminate between fully regenerated (\textbf{FR}) and non-FR images, exploiting explicit FR labels available in the training set. A CLIP-ViT~\cite{clip} backbone is adapted via LoRA~\cite{lora} fine-tuning, with inputs randomly cropped to the model resolution (\(336{\times}336\)) and augmented by horizontal flips. This classifier not only provides accurate FR vs. non-FR discrimination, but also forms the foundation for dense localization in the next stage. A key insight is that a classifier trained with global FR supervision can be repurposed, without retraining, to produce localized heatmaps of synthetic evidence.

\vspace{-5mm}
\subsubsection{Stage 2: Localization via Sliding Classifier Heatmaps}
At inference time, the FR classifier is reused to generate a dense heatmap of synthetic likelihoods. Given an RGB image \(\mathbf{I}\in[0,1]^{3\times H\times W}\), we extract overlapping tiles \(\mathbf{T}_{y,x}\) of size \(p{\times}p\) (e.g., \(336{\times}336\)) with stride \(s\) (e.g., \(112\)), applying reflect padding at borders. Each tile is passed through the classifier to produce an FR logit \(\ell_{y,x}\). To avoid discontinuities at tile boundaries, we weight each prediction by a smooth window \(w\in[0,1]^{p\times p}\) (Hamming/Hann)~\cite{hann-blending} normalized to unit peak. Letting \(\Omega_{y,x}\) denote the spatial support of tile \((y,x)\), we accumulate logits and weights as
\vspace{-1mm}
\begin{align}
S(u,v) &= \sum_{(y,x):\,(u,v)\in\Omega_{y,x}} \ell_{y,x}\; w_{y,x}(u,v),\\
W(u,v) &= \sum_{(y,x):\,(u,v)\in\Omega_{y,x}} w_{y,x}(u,v),
\end{align}
\vspace{-1mm}
and obtain the dense probability map
\vspace{-1mm}
\begin{equation}
\mathbf{H}(u,v)\;=\;\sigma\!\left(\frac{S(u,v)}{\max\{W(u,v),\varepsilon\}}\right),\quad \mathbf{H}\in[0,1],
\end{equation}
where \(\sigma\) is the logistic function and \(\varepsilon\) prevents division by zero. The result is a per-pixel FR heatmap that highlights regenerated content consistently across overlaps and resolutions. 

By transforming a lightweight FR classifier into a dense localization mechanism, the SH branch turns global supervision into pixel-level guidance. Together with the contour-aware EGS branch, it supplies complementary region-level evidence that remains reliable in ambiguous patches and scales gracefully to very large images.


\vspace{-5mm}
\subsection{Two-Branch Fusion Strategy}
\vspace{-3mm}
The two branches are complementary: for small images, the EGS operates at an appropriate scale and tends to be more reliable, whereas for high-resolution images that must be split into many patches, the SH branch provides strong region-level cues.
We fuse them by a pixel-wise max on the probability maps. 
Let \(\mathbf{P}_{\mathrm{EGS}}\in[0,1]^{H\times W}\) denote the EGS probability map and \(\mathbf{P}_{\mathrm{SH}}\in[0,1]^{H\times W}\) the SH heatmap. 
The fused score is
\vspace{-1mm}
\vspace{-1mm}
\vspace{-1mm}
\[
\mathbf{P}_{\mathrm{fuse}}(u,v) \;=\; \max\!\big(\mathbf{P}_{\mathrm{EGS}}(u,v),\, \mathbf{P}_{\mathrm{SH}}(u,v)\big),
\]
which corresponds to a logical OR when maps are binary. 
The final binary mask is obtained by thresholding
$\hat{\mathbf{Y}}(u,v) \;=\; \mathds{1}\!\left[\, \mathbf{P}_{\mathrm{fuse}}(u,v) \ge \tau^\ast \,\right],$
with \(\tau^\ast\) chosen on a validation set.

\vspace{-5mm}
\section{Experiment}\label{sec:experiment}
\vspace{-3mm}

In this section, we first describe our experimental setup and then present extensive experiments to quantify the contribution of each component under cross-domain evaluation, using Intersection over Union (IoU) as the primary metric.

\vspace{-4mm}
\subsection{Experimental Setup}
\vspace{-2mm}

\subsubsection{Dataset}
We follow the Manipulated Region Localization task (Task B) in the MediaEval 2025 SynthIM challenge~\cite{mediaeval-synthim}. The official train/validation split~\cite{mediaeval2025-sid} is used without any external data. 

\vspace{-2mm}
\paragraph{Training.} 
The training set is derived from TGIF~\cite{tgif}, containing about 75k text-guided inpainting samples generated by SD2, SDXL, and Adobe Firefly. Each original image (OR) is used to generate \textbf{spliced} (SP) and \textbf{fully regenerated} (FR) images. For localization, we use only OR and SP images, assigning black masks to OR and segmentation masks to SP, discarding bounding-box annotations. 

\vspace{-2mm}
\paragraph{Validation.} 
Evaluation is performed on the SynthIM validation split, a subset of SAGI-D~\cite{sagi} generated by diverse inpainting models including BrushNet, PowerPaint, HD-Painter, ControlNet, Inpaint-Anything, and Remove-Anything. It contains 9,439 images (3,227 OR, 3,043 SP, 3,169 FR); the test set is withheld by the organizers. We focus on the 3,043 SP images. Unlike TGIF’s fixed resolutions (512–1024px), these span from $\sim$100k pixels to over 100M pixels, making the benchmark a test of cross-domain generalization and robustness to unseen models and extreme scales.

\vspace{-3mm}
\subsubsection{Evaluation Criteria}
Localization performance is measured by Intersection over Union (IoU). Predicted probability maps are binarized at thresholds from \(0.0\) to \(1.0\) (step \(0.01\)), and true positives (TP), false positives (FP), and false negatives (FN) are counted against the ground-truth mask. The IoU is defined as
\vspace{-1mm}
\[
\mathrm{IoU} = \frac{\mathrm{TP}}{\mathrm{TP} + \mathrm{FP} + \mathrm{FN}}.
\]
For \textbf{OR} images, the ground-truth mask is empty, making IoU trivial (always 0 or 1 depending on false positives) and thus uninformative. 
Following common practice, we report mean IoU only on \textbf{SP} images with annotated manipulated regions, averaging over all samples in the validation set.

\vspace{-3mm}
\subsubsection{Implementation Details}

All models are implemented in PyTorch and trained on a single NVIDIA A100 GPU. The system consists of two complementary components: the SH branch and the EGS branch.

\vspace{-3mm}
\paragraph{SH branch.}
We fine-tune a CLIP-ViT Large (14/336)~\cite{clip} using LoRA~\cite{lora} to classify images as fully regenerated (FR) versus non-FR (SP and OR). 
LoRA adapters are inserted into all attention blocks with rank \(r{=}8\), scaling factor \(\alpha{=}32\), and dropout \(0.05\). 
Training uses AdamW (lr \(1{\times}10^{-4}\), weight decay \(1{\times}10^{-4}\)), batch size \(32\), and random \(336{\times}336\) crops with horizontal flips. 
At inference, the classifier produces a dense heatmap by sliding a \(336{\times}336\) window with stride \(112\), and overlapping predictions are merged using Hann blending~\cite{hann-blending}.

\vspace{-3mm}
\paragraph{EGS branch.}
We integrate the pretrained Edge Detector (Section~\ref{sec:edge_detector}) with a SegFormer MiT-B3 decoder~\cite{segformer}. 
Inputs are zero-padded so that both dimensions are divisible by 32; within each batch, all samples share the same padded resolution. 
The Edge Detector is trained once and frozen, providing stable priors fused with RGB for SegFormer. 
Optimization uses Adam (lr \(1{\times}10^{-5}\), weight decay \(1{\times}10^{-4}\)), cosine annealing (\(T_{\max}{=}100\), \(\eta_{\min}{=}1{\times}10^{-6}\)), mixed precision, and batch size 8. 
At test time, large images are divided into non-overlapping tiles (\(\leq1024\) px per side), grouped into padded batches, processed independently, and reassembled. 
This unified padding-and-batching strategy avoids boundary artifacts and ensures consistent train–test behavior.

\vspace{-5mm}
\subsection{Comparison with Existing Localization Methods}
\vspace{-1mm}

We compare our framework with representative manipulation localization models, including IML-ViT~\cite{iml-vit}, MVSSNet~\cite{mvss}, and CATNet~\cite{catnet}. 
All baselines are trained and evaluated under identical settings using the IMDLBenco~\cite{imdlbenco} implementation for fair comparison. 
Table~\ref{tab:quantitative-comparison} reports mean IoU across three evaluation modes: 
\emph{resize} (entire image downscaled), \emph{patch} (per-patch inference), and \emph{merged patch} (stitched full-resolution prediction).  

All models exhibit degraded performance in the \emph{resize} mode due to the loss of fine boundary cues after downscaling. 
Existing methods also show limited robustness when trained on medium-resolution TGIF images but evaluated on the high-resolution, cross-domain SynthIM validation set. 
In contrast, our patch-based design preserves full detail without resizing: the EGS branch achieves 0.442 merged mIoU, while combining it with the SH branch yields the best overall performance of 0.590.

\begin{table}[t]
\centering
\small
\begin{tabular}{lccc}
\toprule
\textbf{Method} & \textbf{Resize IoU} & \textbf{Patch IoU} & \textbf{Merged IoU} \\
\midrule
IML-ViT~\cite{iml-vit}              & 0.301 & 0.531 & 0.362 \\
MVSSNet~\cite{mvss}             & 0.292 & 0.537 & 0.355 \\
CATNet~\cite{catnet}               & 0.285 & 0.528 & 0.351 \\
SegFormer-B3 (baseline)~\cite{segformer} & 0.305 & 0.551 & 0.370 \\
\midrule
Ours (EGS)                         &  --   & 0.597 & 0.442 \\
Ours (EGS + SH)                    &  --   &  --   & \textbf{0.590} \\
\bottomrule
\end{tabular}
\caption{
Quantitative comparison on the SynthIM validation set (mIoU). 
\textbf{EGS}: Edge-Guided Segmentation branch; \textbf{SH}: Synthetic Heatmapping branch. 
``--'' indicates the metric is not applicable. 
}
\label{tab:quantitative-comparison}
\vspace{-2mm}
\end{table}

\subsection{Ablation Study}
\vspace{-2mm}

\begin{table}[t]
\centering
\setlength{\tabcolsep}{10pt}
\begin{tabular}{lcc}
\toprule
\textbf{Method} & \textbf{Patch mIoU} $\uparrow$ & \textbf{Image mIoU} $\uparrow$ \\
\midrule
SegFormer (resized baseline)                 & ---   & 0.305 \\
SegFormer (patching/tiling)                  & 0.551  & 0.370 \\
Edge-Guided SegFormer (patching/tiling)      & 0.597 & 0.442 \\
Synthetic Heatmapping (sliding-window)       & ---   & 0.384 \\
Fusion: EGS + SH            & ---   & \textbf{0.590} \\
\bottomrule
\end{tabular}
\caption{Ablation of pipeline components. \emph{Patching/tiling} denotes non-overlapping patches processed at native scale and stitched back; \emph{Patch mIoU} is measured before stitching, while \emph{Image mIoU} is measured after stitching or blending. 
The \emph{resized baseline} downsamples the entire image to the model input size. 
\emph{Edge-Guided SegFormer} uses the frequency-based edge prior from Section~\ref{sec:edge_detector}. 
Dashes indicate not applicable.}
\label{tab:ablation}
\vspace{-7mm}
\end{table}

\begin{table}[t]
    \centering
    \setlength{\tabcolsep}{10pt}
    \begin{tabular}{lcc}
        \toprule
        \textbf{Edge–Segmentation Integration} & \textbf{Patch mIoU} $\uparrow$ & \textbf{Image mIoU} $\uparrow$ \\
        \midrule
        Edge prior $\rightarrow$ U\!-\!Net (no RGB)                  & 0.498 & 0.366 \\
        Channel adapter (Edge + RGB) $\rightarrow$ SegFormer         & 0.554 & 0.372 \\
        Spatial stacking [RGB; Edge] $\rightarrow$ SegFormer         & 0.588 & 0.438 \\
        Spatial stacking + edge augmentation $\rightarrow$ SegFormer & \textbf{0.597} & \textbf{0.442} \\
        \bottomrule
    \end{tabular}
    \caption{Ablation of edge-segmentation coupling strategies. 
    \emph{Edge prior $\rightarrow$ U\!-\!Net} feeds only the edge map into a U\!-\!Net decoder; 
    \emph{Channel adapter} fuses Edge+RGB via a learned adapter before SegFormer; 
    \emph{Spatial stacking} concatenates RGB and the edge prior as separate spatial bands; 
    \emph{+ edge augmentation} applies light perturbations to the edge prior during training. 
    \emph{Patch mIoU} is computed on tiles before stitching, while \emph{Image mIoU} is measured after stitching.}
    \label{tab:edge_integration_ablation}
\end{table}

\begin{figure}[!!h]
    \vspace{-7mm}
    \centering
    \includegraphics[width=0.9\linewidth]{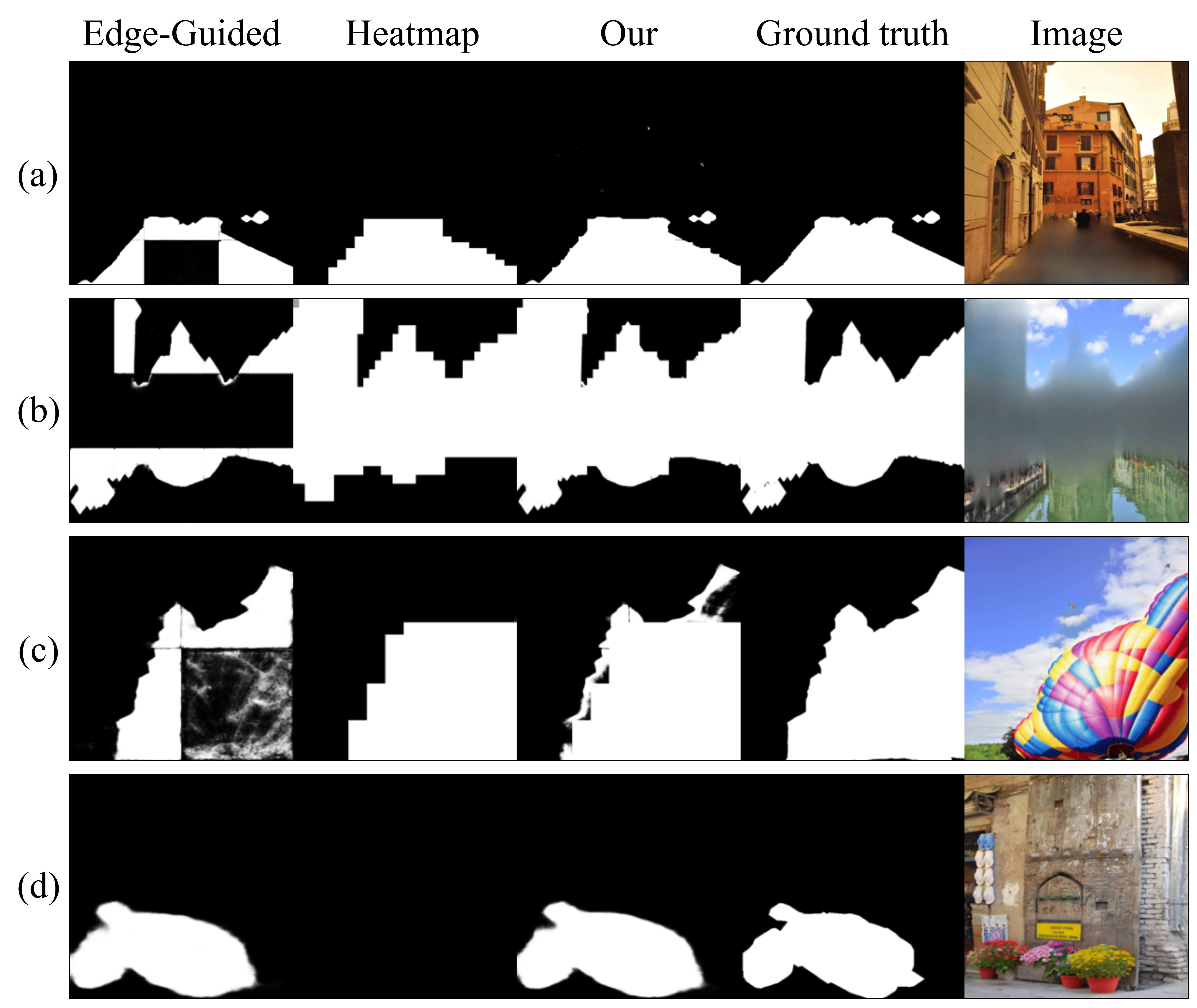}
    \vspace{-5mm}
    \caption{Each group of five images shows, from left to right, the EGS mask, the SH mask, the fuse mask, the ground-truth mask, and the original image. In (a)–(c), showing large images, SH supplies strong region cues that complement EGS. In (d), showing a small image, EGS captures fine details and performs well even without SH. The fused output combines these complementary strengths.}
    \label{fig:edge-heatmap-our-gt-image}
    \vspace{-7mm}
\end{figure}

We conduct ablations to assess the contribution of each component, focusing on the dual-branch design and the role of the edge detector.

As reported in Table~\ref{tab:ablation}, a baseline SegFormer (resized)~\cite{segformer}, which downsamples the whole image to model input size, achieves only 0.305 mIoU due to severe detail loss. 
Switching to a patch-based strategy improves results to 0.370 mIoU at the image level (0.551 patch-level). 
Adding edge guidance from our pretrained detector further sharpens boundaries and improves region consistency, raising performance to 0.597 patch mIoU and 0.442 image mIoU. 
Separately, the SH branch reaches 0.384 mIoU, and fusing SH with EGS yields the best performance of 0.590 mIoU, confirming that the branches provide complementary cues.

To analyze edge integration strategies (Table~\ref{tab:edge_integration_ablation}), we first decode the edge map with a lightweight U-Net~\cite{unet}, which gives 0.366 mIoU. 
Concatenating edge maps with RGB via a \(1\times1\) adapter yields only a minor gain (0.372 mIoU). 
Our spatial stacking approach (Section~\ref{sec:edge_detector}) significantly improves performance to 0.438 mIoU. 
Further applying edge augmentations during training, which simulate degraded or fragmented contours, achieves the best robustness at 0.597 patch mIoU and 0.442 image mIoU. 
These results highlight that spatial stacking with augmented edge priors is the most effective way to integrate geometric cues for manipulation localization.


\begin{figure}[!!h]
    \vspace{-2mm}
    \centering
    \includegraphics[width=1\linewidth]{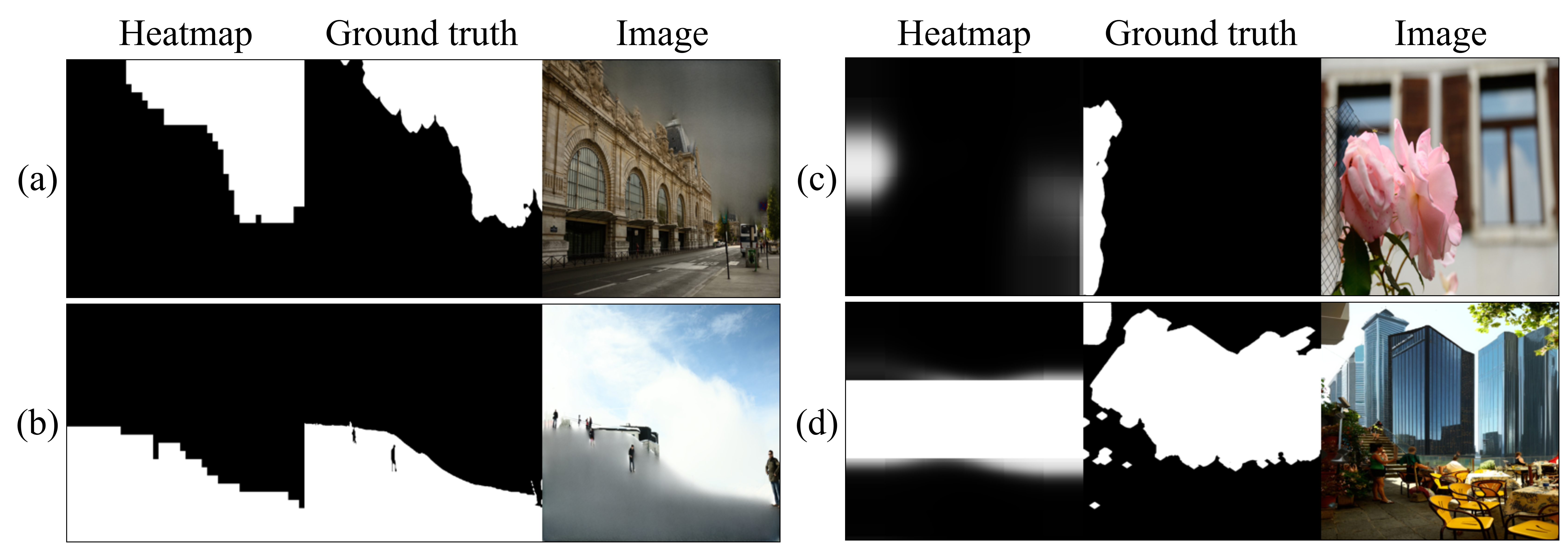}
    \vspace{-9mm}
    \caption{Each triplet shows, from left to right, the SH mask, the ground-truth mask, and the original image. In (a)–(b), showing large images, SH captures coarse regions but misses fine boundaries and small structures. In (c)-(d), showing small images, SH produces blurry responses and overlooks many manipulated pixels. This shows that SH alone is insufficient without complementary edge guidance and fusion.}
    \label{fig:heatmap-gt-image}
    \vspace{-4mm}
\end{figure}

\vspace{-5mm}
\subsection{Qualitative Result}
\vspace{-2mm}

In Fig.\ref{fig:edge-heatmap-our-gt-image}, we qualitatively compare the masks produced by the EGS branch, the SH branch, and their fused prediction on four representative examples. The two cues are complementary: SH yields clean regions for large, smooth structures, whereas EGS preserves thin boundaries and small details. Panels (a)-(c) contain large images where SH corrects EGS regions. In contrast, panel (d) contains a small image for which EGS alone closely matches the ground truth. Overall, the fused mask shows the strengths of both and aligns best with the ground truth.

In Fig.\ref{fig:edge-heatmap-our-gt-image}, we observed that the SH often produces visually plausible regions. A natural question is whether the SH branch alone is sufficient. Fig.\ref{fig:heatmap-gt-image} examines this. In panels (a)–(b), showing large images, SH yields smooth, coarse masks but fails to preserve thin structures and boundary details. In panels (c)–(d), showing small images, SH responses become diffuse and omit many manipulated pixels. These limitations motivate the need for the edge-guided stream and the fusion module. This aligns with the quantitative results in Table~\ref{tab:ablation}, where SH alone reaches 0.384 image mIoU, notably lower than the fusion model.

\vspace{-3mm}
\section{Conclusion}\label{sec:conclusion}
\vspace{-3mm}

Existing IFL models often struggle to generalize across real-world conditions.
Most of them are trained on fixed input resolutions and rely heavily on semantic content, making them less effective on large or unseen images where manipulations vary in scale and texture.
To address these issues, we proposed EDGER, a dual-branch, patch-based framework that unifies fine-grained boundary localization with region-level reasoning.
The EGS branch employs a Frequency-based Edge Detector to provide stable, non-semantic boundary cues for a SegFormer backbone, while the SH branch leverages a CLIP-ViT classifier to capture region differences through sliding-window inference. 
Fusing their outputs by a pixel-wise max produces accurate and consistent masks over arbitrary resolution images.
Experiments on the validation set show that our method improves both boundary precision and region accuracy, outperforming conventional SegFormer and other state-of-the-art baselines. 
Future work will focus on adaptive branch fusion and end-to-end joint optimization to further enhance generalization to in-the-wild manipulated images.

\section*{Acknowledgments}
This research is funded by Vietnam National University, Ho Chi Minh City (VNU-HCM) under grant number DS.C2025-18-13


\bibliographystyle{splncs04}
\bibliography{references}
\end{document}